\documentclass[10pt,twocolumn,letterpaper]{article}

\usepackage{cvpr}
\usepackage{times}
\usepackage{epsfig}
\usepackage{graphicx}
\usepackage{amsmath}
\usepackage{amssymb}
\usepackage{comment}
\usepackage{multirow}
\usepackage{colortbl}
\usepackage[table]{xcolor}
\usepackage[marginal]{footmisc}
\usepackage{comment}

\usepackage[breaklinks=true,bookmarks=false]{hyperref}

\cvprfinalcopy % *** Uncomment this line for the final submission

\def\cvprPaperID{1288} % *** Enter the CVPR Paper ID here

\ifcvprfinal\fi

\newcommand\RedeclareMathOperator{%
	\@ifstar{\def\rmo@s{m}\rmo@redeclare}{\def\rmo@s{o}\rmo@redeclare}%
}

\DeclareMathOperator{\ep}{ep}
\DeclareMathOperator{\EP}{EP}

\DeclareMathOperator{\bigO}{O}
\DeclareMathOperator{\LSTM}{LSTM}
\DeclareMathOperator{\softmax}{softmax}
\DeclareMathOperator{\mif}{if}

\newcommand{\defeq}{\stackrel{\text{def}}{=}}

\begin{document}

%%%%%%%%% TITLE
\title{Edit Probability for Scene Text Recognition}

\begin{comment}
\author{Fan Bai\\
Fudan University\\
{\tt\small fbai17@fudan.edu.cn}
\and
Zhanzhan Cheng\\
Hikvision Research\\
{\tt\small chengzhanzhan@hikvision.com}
\and
Yi  Niu\\
Hikvision Research\\
{\tt\small niuyi@hikvision.com}
\and
Shiliang  Pu\\
Hikvision Research\\
{\tt\small pushiliang@hikvision.com}
\and
Shuigeng  Zhou\thanks{Corresponding author.}\\
Fudan University\\
{\tt\small sgzhou@fudan.edu.cn}
}
\end{comment}

\author{Fan Bai\textsuperscript{1}\thanks{Fan Bai did most of this work when he was an intern in Hikvision Research Institute.}\qquad\qquad
Zhanzhan Cheng\textsuperscript{2}\qquad\qquad
Yi Niu\textsuperscript{2} \\
Shiliang Pu\textsuperscript{2}\qquad\qquad\qquad
Shuigeng Zhou\textsuperscript{1}\thanks{Corresponding author.}\\
\textsuperscript{1}Shanghai Key Lab of Intelligent Information Processing, and School of \\Computer Science, Fudan University, Shanghai 200433, China\\
\textsuperscript{2}Hikvision Research Institute, China\\
{\tt\small \{fbai17,sgzhou\}@fudan.edu.cn}\\
{\tt\small \{chengzhanzhan,niuyi,pushiliang\}@hikvision.com}
%\and
%Fan Bai\\
%Fudan University\\
}

\maketitle

%%%%%%%%% ABSTRACT
%\begin{abstract}
%% background
%    Sequence recognition has been a lasting hot research topic in sequence learning, and the most popular framework is attention-based encoder-decoder framework.
%    In this paper, we investigate the problem of scene text recognition, which is among the most challenging tasks in sequence recognition.
%    In scene text recognition, the encoder-decoder-based approach is the-state-of-the-art.
%% defect of existing method
%    However, we observe that existing objective functions is not suitable for this framework, because some incorrect insertions or deletions may occur in the output sequence of probability distribution.
%    For example, framewise classification, such as cross entropy (CE) does not take into account contextual information,
%    % and connectionist temporal classification (CTC) only considers unexpected insertions by skipping some temporal distribution,
%    which  may confuse the whole model learning process due to some dislocations in back-propagation.
%% new method
%    Instead of CE, we propose a novel method \textbf{e}dit \textbf{p}robability (EP) for text recognition, which considers both insertion and deletion problems in training.
%% experiment
%    Extensive experiments on standard benchmarks, including the IIIT-5K, Street View Text and ICDAR datasets, show that the EP-based method substantially outperforms the existing methods.
%\end{abstract}
\begin{abstract}
	% background
	We consider the scene text recognition problem under the attention-based encoder-decoder framework, which is the state of the art.
	%    In this paper, we investigate the problem of scene text recognition, which is among the most challenging tasks in sequence recognition.
	%    In scene text recognition, the encoder-decoder-based approach is the-state-of-the-art.
	% defect of existing method
	The existing methods usually employ a frame-wise maximal likelihood loss to optimize the models.
	When we train the model, the misalignment between the ground truth strings and the attention's output sequences of probability distribution, which is caused by missing or superfluous characters, will confuse and mislead the training process, and consequently make the training costly and degrade the recognition accuracy.
	%    However, we observe that existing objective functions is not suitable for this framework, because some incorrect insertions or deletions may occur in the output sequence of probability distribution.
	%    For example, framewise classification, such as cross entropy (CE) does not take into account contextual information,
	% and connectionist temporal classification (CTC) only considers unexpected insertions by skipping some temporal distribution,
	%    which  may confuse the whole model learning process due to some dislocations in back-propagation.
	% new method
	To handle this problem, we propose a novel method called edit probability (EP) for scene text recognition. EP tries to effectively estimate the probability of generating a string from the output sequence of probability distribution conditioned on the input image, while considering the possible occurrences of missing/superfluous characters.
	The advantage lies in that the training process can focus on the missing, superfluous and unrecognized characters, and thus the impact of the misalignment problem can be alleviated or even overcome.
	%    Instead of CE, we propose a novel method \textbf{e}dit \textbf{p}robability (EP) for text recognition, which considers both insertion and deletion problems in training.
	% experiment
	We conduct extensive experiments on standard benchmarks, including the IIIT-5K, Street View Text and ICDAR datasets. Experimental results show that the EP can substantially boost scene text recognition performance.
\end{abstract}

\section{Introduction}\label{sec:intr}
% sequence learning background
Text recognition in natural scene images has recently attracted much research interest of the computer vision community~\cite{jaderberg2014synthetic,lee16recursive,shi17end}.
%And the state of the art achieves substantial performance improvement by applying sequence recognition framework\cite{neumann2012real, jaderberg2014synthetic, 7801919}[].
The sequence-learning-based text recognition techniques, which have been advancing rapidly in recent years~\cite{cheng17fan,lee16recursive,shi2016robust},
%A Sequence-learning-based text recognizer
generally consist of an encoding module and a decoding module.
% encoding module and decoding module
%Concretely,
The encoding module usually encodes the input images to vectors of fixed dimensionality with a certain encoding technique, such as convolution neural network (CNN) \cite{shi2016robust}, or recurrent neural network (RNN) including long short-term memory (LSTM)~\cite{hochreiter97lstm,shi17end,sutskever14seq} and gate recurrent neural network (GRU)~\cite{bahdanau2016end,cho14gru,shi2016robust}.
While the decoding module decodes the encoded feature vectors into the target strings by exploiting RNN, connectionist temporal classification (CTC)~\cite{alex06ctc,shi17end} or attention mechanism~ \cite{bahdanau15neural,bahdanau16end,shi2016robust} \etc.

% encoder-decoder background in scene text reading
%According to previous works\cite{shi2016robust, 7801919,Lee_2016_CVPR},
The state-of-the-art of scene text recognition is the attention-based encoder-decoder framework~\cite{cheng17fan,lee16recursive,shi2016robust}. It outputs a sequence of \emph{probability distribution} (\emph{pd}) that is expected to be aligned with the characters of the text in the input image.
%Thus, it is generally trained to maximize the probability of each \emph{gt}-character in the corresponding output \emph{pd} sequence with a frame-wise maximal likelihood loss function.
In model training, the probability of the \emph{ground-truth text} (\emph{gt} in short), calculated by the corresponding output \emph{pd} sequence in a frame-wise style, is utilized to estimate the likelihood of model parameters. In this paper, we call this joint probability \emph{frame-wise probability} (FP), and the existing frame-wise loss based methods \emph{FP based methods} in the sequel.
However, in both the training and predicting processes of the attention-based text recognition models, some characters may be missing or superfluous, which results in misalignment between the \emph{gt} and the output sequence of \emph{pd}.
In a recent work, Cheng \etal~\cite{cheng17fan} considered this phenomenon as a result of attention drift, and solved it by introducing the Focusing Attention Net~(FAN). This method achieved the state of the art performance.
But the training of FAN requires extra pixel-wise supervising information, which is expensive to provide. And the training process is time-consuming due to the large amount of pixel-wise calculation.
%The misalignment problem has also been observed in attention training of speech recognition. And Kim \etal blame the misalignment issues to the lack of left-to-right constraints as used in CTC, and overcome that by using a joint CTC-attention model within the multi-task learning framework[]. However, Cheng \etal point out that the joint CTC-attention model does not work well in scene text reading\cite{2017arXiv170902054C}.

\begin{figure}[!thbp]
	\begin{center}
		\includegraphics[width=0.45\textwidth]{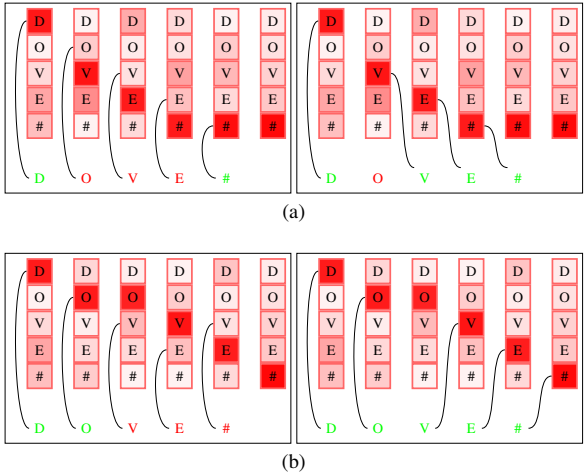}
	\end{center}
	\caption{The illustration of misalignment between the output sequences of \emph{pd} and the ground truth ``DOVE\#'' due to (a) missing character and (b) superfluous character.
		%The deeper red of a block refers to the higher probability of the character in the block.
		A cell with higher saturation corresponds to a character with higher probability in the $pd$.
		The green \emph{gt}-characters at the bottom of each subfigure have the highest probability in their \emph{pd}s (linked by curves), while the red characters have lower probability in their $pd$s.
		 %, bellow in each sub-figure.
		In subfigure (a), the left part shows that `O', `V' and `E' have low probability in their corresponding \emph{pd}s. The right part shows if an `O' is inserted, then `V' and `E' are alignable with the probabilities in their $pd$s.
		Similarly, in subfigure (b), if an 'O' is removed, then the remaining `O', `V', `E' and `\#' have matched probabilities in their $pd$s.
		%The left figures of (a) and (b) separately represent the incorrect deletion and incorrect insertion yielded with cross entropy, while the right parts show the better way to generate the expected target sequence.  For each unit in the probability distribution sequence,  the deeper red represents a higher probability, vice versa.
		% Fig (a)
		%In the left figure of (a),  the frame-wise classification result over the probability distribution sequence is ``ACD\#\#'', and the `B' is missing;
		%Then the cross entropy loss is calculated with aiming to maximize the likelihood of the ground truth conditioned on the probability distribution sequence column by column, while a better way (shown in the right figure) is to insert a `B' between the 1st and 2nd columns, which is more low-cost for back-propagation.
		%Similarly,  the left figure of (b) shows that the corresponding classification result is ``ABBCD\#'', and a `B' is superfluous; In back-propagation stage, a better manner is to delete one of duplicated `B' columns for easily learning.
		%The green (red) characters represent that the probability of the expected characters is (is not) dominant in columns linked with the  corresponding  lines.
	}
	\label{fig:mot}
\end{figure}

Fig.~\ref{fig:mot} provides examples to illustrate the phenomenon of missing and superfluous characters in training an attention-based text recognition model on the ground truth ``DOVE\#''. Here, `\#' represents the End-Of-Sequence (EOS) symbol, which is commonly used in attention-based methods~\cite{cheng17fan,lee16recursive,shi2016robust}.
In Fig.~\ref{fig:mot} (a) and (b), the model may recognize the inputs as ``DVE\#'' and ``DOOVE\#'' respectively, based on the output $pd$ sequences. Comparing against the $gt$ ``DOVE\#'', it is natural to say that the former misses an `O' and the latter has a superfluous `O'.

By checking the training process of attention-based text recognition models, we can see that the FP-based methods simply train the model by maximizing the probability of each character in the \emph{gt}, based on the corresponding \emph{pd} of the attention's output sequence. %, but completely unconcerns about whether they are correctly aligned(\eg left parts of Fig.~\ref{fig:mot}).
However, the misalignments caused by missing or superfluous characters may confuse and mislead the training process, and consequently make the training costly and degrade the recognition accuracy.
Concretely, back to Fig.~\ref{fig:mot}, except for `D' and `\#', all the other three characters `O', `V' and `E' have low probability in the corresponding \emph{pd}s (see the left diagrams), thus large error will be back-propagated to these \emph{pd}s for further training.
Instead, if the training algorithm realizes that the character `O' in Fig.~\ref{fig:mot} (a) is missing and one of the character `O' in Fig.~\ref{fig:mot} (b) is superfluous, it will align `V' and `E' in Fig.~\ref{fig:mot} (a), `V', `E', and `\#' in Fig.~\ref{fig:mot} (b) to more appropriate \emph{pd}s (see the right diagrams), and the characters will have higher probability under the new alignment, then it needs only to focus on the missing/superfluous character `O', which will make the training simpler and less costly.

Motivated by the observation above and inspired by the concept of sequence alignment where edit distance is used to measure the dissimilarity of two sequences, in this paper we propose a new method for scene text recognition under the attention-based encoder-decoder framework, which is called \textbf{edit probability} (\textbf{EP} in short).
By treating the misalignment between the $gt$ and the output $pd$ sequence as the result of possible occurrences of missing/superfluous characters, in the training process, EP tries to effectively estimate the probability of a string conditioned on the output sequence of \emph{pd} under certain model parameters while considering possible occurrences of missing/superfluous characters.
The merit lies in that the training process can focus on the missing, superfluous and misclassified characters, and the impact of misalignment can be reduced substantially.
To validate the proposed method, we conduct extensive experiments on several benchmarks, which show that EP can significantly boost recognition performance.

The rest of this paper is organized as follows: Section~\ref{sec:related-work} reviews related work, Section~\ref{sec:method} presents the EP method in detail, Section~\ref{sec:performance} conducts empirical evaluation, and Section~\ref{sec:conclusion} concludes the paper.

%------------------------------------------------------------------------
\section{Related Work}\label{sec:related-work}
%------------------------------------------------------------------------
% current loss
%At the present stage, most of encoder-decoder frameworks are trained with Cross Entropy Loss [] or CTC Loss.
%However, both of CE and CTC can not tackle with incorrect insertions and deletions in the output probability distribution sequence well.
% CE loss
%The CE calculates cross entropy between the sequence of ground truth and the output sequence column by column, but it does not take into account contextual relationships.
% CTC loss
%The Connectionist Temporal Classification is another popular technique for mapping the encoded sequences to target sequences, but it only considers incorrect insertions.
% EP
%In order to consider both insertions and deletions in sequence generation, we design the EP to capture the target deleting or inserting probability.

%In scene text recognition tasks, the encoder-decoder-based sequence recognition framework has been wildly used in previous works[][][][][].
%We find that most of encoder-decoder frameworks are trained with Cross Entropy Loss [] or CTC Loss.
%The CE-based methods calculates cross entropy as the loss between the sequence of ground truth and the output sequence column by column, but it does not take into account contextual relationships. While the CTC-based methods calculate loss by mapping the encoded sequences to target sequences, but it only considers incorrect insertions.

% background
In the past decade, many methods have been proposed for scene text recognition, which roughly fall into three types: 1) traditional methods with handcrafted features, 2) Na\"ive deep neural-network-based methods, and 3) sequence-based methods.

% traditional methods
In early years, traditional methods first extract handcrafted visual features for individual character detection and recognition one by one, then integrate these characters into words based on a set of heuristic rules or a language model.
Neumann and Matas~\cite{neumann2012real} recognized characters by training a Support Vector Machine (SVM) classifier with the defined handcrafted features such as aspect ratio, hole area ratio etc. Wang \etal~\cite{wang2011end,wang2010word} first trained a character classifier with the extracted HOG descriptors~\cite{yao2014strokelets}, then recognized characters of the cropped word image by a sliding window one by one.
However, due to the low representation capability of handcrafted features, these methods cannot achieve satisfactory recognition performance.

% cnn-based methods
Later, instead of handcrafted features, some deep neural-network-based methods were developed for extracting robust visual features.
Bissacco \etal~\cite{bissacco2013photoocr} adopted a fully connected network (FC) of 5 hidden layers for extracting character features, then applied an n-gram language model to recognize characters.
Wang \etal~\cite{wang2012end} and Jaderberg \etal~\cite{jaderberg2014deep, jaderberg2016reading} first developed CNN-based framework for character feature representation, then applied some heuristic rules for characters generation.
These na\"ive deep neural-network-based methods above usually recognized character sequence based on some pre/post-processing, such as the segmentation of each character or a non-maximum suppression, which may be very challenging because of the complicated background and the inadequate distance between consecutive characters.

% sequence-based methods
Recently, some researchers treated the text recognition task as a sequence learning problem: first encoding a text image into a sequence of features with deep neural network, then directly generating character sequence with sequence recognition techniques.
He \etal~\cite{he2016deep} and Shi~\etal\cite{shi17end} proposed the end-to-end neural networks that first capture visual feature representation by using CNN or RNN, then the CTC~\cite{alex06ctc} loss was combined with the neural network outputs for calculating the conditional probability between the predicted and the target sequences.
The state of the art for text recognition is the attention-based methods~\cite{cheng17fan,lee16recursive,shi2016robust}. %Yang \etal [],
%A latest such method was proposed by Cheng \etal~\cite{2017arXiv170902054C}.
These methods first combined CNN and RNN for encoding text images into feature representations, then employed a frame-wise loss to optimize the model. In the training process, the misalignment between the $gt$ sequence and the output $pd$ sequence may mislead the training algorithm and result in poor performance.

Note that the misalignment problem has also been observed in attention training of speech recognition. Kim \etal
%attributed the misalignment problem to the lack of left-to-right constraints as used in CTC, which is designed , and
tried to solve the problem by using a joint CTC-attention model within the multi-task learning framework~\cite{kim2016joint}. However, as pointed in \cite{cheng17fan}, the joint CTC-attention model does not work well in scene text recognition.
%And it is theoretically unfeasible to attach a CTC decoder to the attention decoder by using the output of attention as the input of CTC, because the CTC decoding requires blanks between characters, which is not supported by attention model.
This paper also addresses the scene text recognition problem under the attention-based framework. Different from the existing methods, we propose a novel method EP that tries to estimate the probability of a string conditioned on the
%output \emph{pd} sequence
input image, by treating the misalignment between the $gt$ text and the output $pd$ sequence as the result of possible occurrences of missing/superfluous characters. EP provides an effective way to handle the misalignment problem, and empirically it outperforms the existing methods.
%{\color{red}give some words here to clarify the difference between EP and the CTC-attention model.}

\section{The EP Method}\label{sec:method}
%In scene text recognition, the attention-based encoder-decoder frameworks are the state of the art. However, existing frame-wise-loss-based attention methods is incapable for learning dislocation problem, which may confuse the model learning process.

In this section, we present the EP method in detail, including the EP-based attention decoder, the formulation of EP, the EP training process, and EP based prediction with/without a lexicon.

\emph{Edit probability} is proposed to effectively train attention-based models for accurate scene text recognition.
Conceptually, for an image $\mathcal{I}$ and a text string $T$, $EP(T|\mathcal{I};\theta)$ measures the probability of $T$ conditioned on $\mathcal{I}$ under model parameters $\theta$. It is evaluated by summing the probabilities of all possible edit paths that transform an initially empty string to $T$ based on the \emph{pd} sequence $y$ generated by the model. %from $\mathcal{I}$.
And each edit path consists of a sequence of edit operations that are detailed in Sec. \ref{sec:ep}.

\subsection{EP-based Attention Decoder}\label{sec:att}
The original attention decoder is an RNN that generates the output \emph{pd} $y_j$ on the $j$-\emph{th} step~\cite{bahdanau2016end}:
\begingroup
\renewcommand*{\arraystretch}{1.4}
\begin{equation}\label{eq:att}
\begin{array}{rcl}
y_j &=& \softmax(v^\top s_j),\\
s_j &=& \LSTM(y_{j-1}, c_j, s_{j-1}),\\
c_j &=& \sum_{k^\prime=1}^{|h|}\alpha_{j,k}h_k,\\
\alpha_{j,k} &=& \dfrac{\exp(e_{j,k})}{\sum_{k^\prime=1}^{|h|}\exp(e_{j,k})},\\
e_{j,k} &=& w^\top\tanh(Ws_{j-1}+Vh_{k}+b)
\end{array}
\end{equation}
\endgroup
where $h$ is a sequence of encoded feature vectors, and $s_j$, $c_j$, $\alpha_j$ and $e_j$ represent the LSTM~\cite{hochreiter97lstm} hidden state, the weighted sum of $h$, the attention weights and the alignment model on the $j$-\emph{th} step, respectively.
$w$, $W$, $V$, $b$ and $v$ are all trainable parameters.
%$g_A$ and $f$ represent a feed-forward network and an LSTM recurrent network, respectively.

In this work, for EP calculation, the attention decoder also generates $R_j$ and $I_j$ on the $j$-\emph{th} step:
\begingroup
\renewcommand*{\arraystretch}{1.25}
\begin{equation}
\begin{array}{rcl}
R_j ~~=~~ (R_j^C, R_j^I, R_j^D)^\top &=& \softmax(W_R^\top s_j),\\
I_j &=& \softmax(W_I^\top s_j)
\end{array}
\end{equation}
\endgroup
where $W_R$ and $W_I$ are trainable parameters.
$R_j^C$, $R_j^I$ and $R_j^D$ respectively represents the probability of \textit{$y_j$ being correctly aligned}, \textit{a character being missing before $y_j$} and \textit{$y_j$ being superfluous}. And $I_j$ is the \emph{pd} of characters being missing before $y_j$ conditioned on $R_j^I$.
%An extra step of attention is required to generate $I_{|A|+1}$ for Eq.~\ref{eq:pi1}.

\subsection{Edit Probability}\label{sec:ep}
% definition
%Edit probability $\ep(T_{1:i} , A_{1:j})$, a mapping from $(\Sigma^*, \mathbb{R}^{|\Sigma|\times L})$ to $ [0,1]$ is the probability of generating the string prefix  $T_{1:i} \in\Sigma^*$ from the first $j$ distributions of the output sequence of probability distribution $A \in \mathbb{R}^{|\Sigma|\times L}$, where $L$ is the length of the probability distribution sequence and $T$  consists of symbols in the category set $\Sigma$.
%And the edit probability is calculated with the three types of edit operation: insertion, deletion and consumption considered.
%There are three types of operation in EP computation: insertion, deletion and consumption.
%And insertion, deletion and consumption are described in Sec. \ref{sec:intr}.
 %, which include 1) appending a character (including the string end symbol EOS) to $S$ directly, 2) deleting the first one from the rest of the output sequence of \emph{pd} directly (if the rest is not empty), and 3) appending a character (or EOS) to $S$ by consuming the first one from the rest of the output sequence of \emph{pd} (if the rest is not empty).

%As shown in Fig.~\ref{fig:ep}, to formally describe the calculation of EP,
Formally, with the alphabet (including the EOS) $\Sigma$, let $T\in\mathbb{T}_\Sigma$
%and $y\in\mathbb{P}_\Sigma^{*}$
where  $\mathbb{T}_\Sigma$
%and $\mathbb{P}_\Sigma$
represents the set of all valid strings (each of which contains one and only one EOS as its end token) on $\Sigma$.
%, and the set of all \emph{pd} on $\Sigma$, respectively.
We define the edit states as tuples $(T_{1:i}, y_{1:j})$ for $0\le i\le|T|$ and $0\le j\le|y|$.
And the state $(T_{1:i}, y_{1:j})$ indicates generating $T_{1:i}$ from $y_{1:j}$.
In particular, $T_{1:0}$ and $y_{1:0}$ represent an empty string (also denoted by ``'') and an empty \emph{pd} sequence respectively.
%By concerning the probability of the occurrence of missing/superfluous characters, the edit operations for state $(T_{1:i}, A_{1:j})$ are defined as follows:
%With $(R_j^C, R_j^I, R_j^D)\in softmax(\mathbb{R}^3)$ and $I_j\in softmax(\mathbb{R}^{|\Sigma|})$ being computed from the LSTM hidden state of attention at $j$-\emph{th} step,
The edit operations for state $(T_{1:i}, y_{1:j})$ are defined as follows:
\begin{itemize}
	%\item \emph{insertion}: an insertion operation over state $(T_{1:i}, A_{1:j})$ appends $T_{i+1}$ to $T_{1:i}$, and transforms state $(T_{1:i}, A_{1:j})$ to state $(T_{1:i+1}, y_{1:j})$ if $i<|T|$. Formally,
	\item \emph{consumption}: the consumption operation $\varepsilon^C_{i,j}$ transforms state $(T_{1:i-1}, y_{1:j-1})$ to state $(T_{1:i}, y_{1:j})$ if $0<i\le|T|$ and $0<j\le|y|$ by regarding the character $T_{i}$ and the \emph{pd} $A_{j}$ as being correctly aligned, and
	appending $T_{i}$ to $T_{1:i-1}$ by consuming $y_{j}$. Formally,
	\begin{equation}
	\varepsilon^C_{i,j}(T_{1:i-1},y_{1:j-1})=(T_{1:i},y_{1:j})\text{.}
	\end{equation}
	The probability of this operation is the joint probability of \textit{doing consumption} and \textit{consuming $T_{i}$ of $y_{j}$}.
	% which respectively denoted by $R_{j+1}^C\in \mathbb{R}$ and $y_{j+1}(T_{i+1})$.
	That is,
	\begin{equation}\label{eq:pc}
	p(\varepsilon^C_{i,j}|\mathcal{I};\theta)=R_{j}^Cy_{j}(T_{i})\text{.}
	\end{equation}
	
	\item \emph{deletion}: the deletion operation $\varepsilon^D_{i,j}$ transforms state $(T_{1:i}, y_{1:j-1})$ to state $(T_{1:i}, y_{1:j})$ if $0<j\le|y|$ by regarding the \emph{pd} $y_{j}$ as being superfluous, and deleting $y_{j}$ directly. Formally,
	\begin{equation}
	\varepsilon^D_{i,j}(T_{1:i},y_{1:j-1})=(T_{1:i},y_{1:j})\text{,}
	\end{equation}
	For $T_i\ne\text{`\#'}$, the probability of this operation is $R_{j}^D$. And for $T_{i}=\text{`\#'}$, \emph{deletion} is the only allowed operation on state $(T_{1:i}, y_{1:j-1})$ as any character after the EOS is ignored. So we have
	\begin{equation}\label{eq:pd}
	p(\varepsilon^D_{i,j}|\mathcal{I};\theta) = \left\{
	\begin{array}{ll}
		 R_{j}^D~~~&\mif T_i\ne\text{`\#'}\text{,} \\
		 1~~~&\mif T_i=\text{`\#'}\text{.}
	\end{array}\right.
	\end{equation}
	
	\item \emph{insertion}: the insertion operation $\varepsilon^I_{i,j}$ transforms state $(T_{1:i-1}, y_{1:j})$ to state $(T_{1:i}, y_{1:j})$ if $0<i\le|T|$ by regarding $T_{i}$ as a missing character, and appending $T_{i}$ to $T_{1:i-1}$ directly. Formally,
	\begin{equation}
	\varepsilon^I_{i,j}(T_{1:i-1}, y_{1:j})=(T_{1:i}, y_{1:j})\text{.}
	\end{equation}%. And the conditional probability of this operation %is the joint conditional probability of doing insertion and inserting the character $T_{i+1}$ before $A_{j+1}$, that is,
	For $j<|y|$, the probability of this operation is the joint probability of \textit{a character is missed from the position just before $y_{j+1}$} and \textit{the missing character is $T_{i+1}$}. And for $j=|y|$, \emph{insertion} is the only allowed operation over state $(T_{1:i-1},y_{1:|y|})$ as there is no more \emph{pd} to delete or consume. So we have
	\begin{equation}\label{eq:pi}
	p(\varepsilon^I_{i,j}|\mathcal{I};\theta) = \left\{
	\begin{array}{ll}
	R_{j}^II_{j}(T_{i})&\mif j<|y|\text{,}\\
	I_{j}(T_{i})&\mif j=|y|\text{.}
	\end{array}\right.
	\end{equation}
\end{itemize}
%In Eq.~\ref{eq:pc}, \ref{eq:pd}, \ref{eq:pd1}, \ref{eq:pi} and \ref{eq:pi1},
%: 1) $R_j(1)$, $R_j(2)$ and $R_j(3)$ represent the probabilities that a character misses just before $A_j$, $A_j$ is superfluous, and $A_j$ is normally recognized, respectively; 2) $I_j\in\mathbb{R}^{|\Sigma|}$ represents the probability distribution of the characters (including EOS) to be inserted if the insertions are performed. While
%$R_j\in\mathbb{R}^3$ and $I_j\in\mathbb{R}^{|\Sigma|}$ are computed from the LSTM hidden state of attention, which will be described in Sec.~\ref{sec:att}.

By assuming that the edit operations over $T$ and $y$ are conditional independent, the probability of an edit path $E$ is the joint probability of all the edit operations in $E$. That is,
\begin{equation}\label{eq:phie}
p(E|\mathcal{I};\theta)=\prod_{t=1}^{|E|} p(E_t|\mathcal{I};\theta),
\end{equation}
where $E_t$ refers to the $t$-\emph{th} edit operation in $E$.
In particular, the probability of an empty path is $1$.

The edit probability of states $(T_{1:i}, y_{1:j})$ is evaluated by the sum of all the conditional probabilities of edit paths $E_{1:|E|}\in\mathbb{E}_\Sigma^*$ that transform $(\text{``''}, y_{1:0})$ to $(T_{1:i}, y_{1:j})$ where ``'' and $y_{1:0}$ represent an empty string and an empty \emph{pd} sequence respectively. Formally,
\begin{equation}\label{eq:ep}
\ep(T_{1:i}, y_{1:j})=\sum_{\substack{E\in\mathbb{E}_\Sigma^*\\E(\text{``''},y_{1:0})=(T_{1:i}, y_{1:j})}}p(E|\mathcal{I};\theta)
\end{equation}
where and $\mathbb{E}_\Sigma$ is the set of all edit operations over $T$ and $y$. That is,
\begin{equation}_{}\label{eq:secep}
%\begin{array}{rcl}
\mathbb{E}_\Sigma \defeq \{\varepsilon^C_{i,j}, \varepsilon^D_{i,j}, \varepsilon^I_{i,j} | 0<i\le|T|, 0< j\le|y|\}\text{.}
%\end{array}\text{.}
\end{equation}
\begin{figure}[!thbp]
	\begin{center}
		\includegraphics[width=0.45\textwidth]{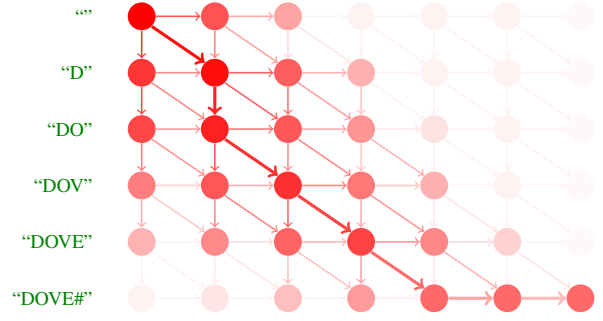}
	\end{center}
	\caption{The illustration of edit probability calculation.
		The red filled circle on the $i$-th row $j$-th column (count from 0) indicates the state $(T_{1:i}, y_{1:j})$,
		in which $T_{1:i}$ is the string prefix (as shown with the green words) and $y_{1:j}$ is the prefix of the \emph{pd} sequence (as shown in Fig.~\ref{fig:mot} (a)).
		Given the state $(T_{1:i}, y_{1:j})$: 1) the horizontal arrow below the state indicates the deletion operation $\varepsilon^D_{i,j}$;
		2) The vertical arrow right to the state indicates the insertion operation $\varepsilon^I_{i,j}$;
		3) The diagonal arrow right below the state indicates the consumption operation $\varepsilon^C_{i,j}$.
		And the higher saturation of a red filled circle or an arrow refers to the higher probability of the state or the corresponding edit operation, vice versa.
		The edit path with the highest probability is emphasized with bold arrows, which consumes $y_{1}$, $y_{2}$, $y_{3}$ and $y_{4}$ to generate `D', `V', `E' and `\#' respectively, inserts an `O' before $y_{2}$, and deletes $y_{5}$ and $y_{6}$.
	}
	\label{fig:ep}
\end{figure}

However, enumerating all the possible edit paths is prohibitively expensive (if not impossible) as the search space is too large.
Fortunately, this can be solved by dynamic programming based on the following equation inferred from Eq.~\ref{eq:phie}:
        \begin{equation}\label{eq:phi1}
		        p(\varepsilon \circ E|\mathcal{I};\theta) = p(\varepsilon|\mathcal{I};\theta)p(E|\mathcal{I};\theta)\text{,}
        \end{equation}
%{\color{red}because performing $E_1$ and $\varepsilon$ is independent}
for $\varepsilon\in\mathbb{E}_\Sigma$ and $E\in\mathbb{E}_\Sigma^*$. And $\circ$ represents the concatenation operator for edit operations and edit paths, which is defined as follows:
\begin{equation}
	(\varepsilon \circ E)(T_{1:i},y_{1:j})=E(\varepsilon(T_{1:i},y_{1:j}))\text{.}
\end{equation}
%Similar to the forward-backward algorithm in CTC~\cite{alex06ctc}, the core idea is that the edit probability of a certain edit state can be computed from the edit probabilities of some other states that can be transformed to the former state with one edit operation.
With Eq.~\ref{eq:phi1}, we can rewrite Eq.~\ref{eq:ep} as follows:
\begingroup
\renewcommand*{\arraystretch}{1.24}
        \begin{equation}\label{eq:eprewrite}
            \begin{array}{rl}
                &\ep(T_{1:i},y_{1:j})\\
%                =&\sum\limits_{\substack{E^\prime \in \mathbb{E}_{\Sigma}^*, \varepsilon \in \mathbb{E}_{\Sigma}  \\ \varepsilon(E^\prime(\text{``''},y_{1:0}))=(T_{1:i},A_{1:j})}} p(E^\prime|\mathcal{I};\theta)p(\varepsilon|\mathcal{I};\theta) \\
%               =& \sum\limits_{\substack{\varepsilon \in \mathbb{E}_\Sigma  \\   (T^\prime, A^\prime) \in (\mathbb{T}_\Sigma, \mathbb{P}_{\Sigma}^*) \\ \varepsilon(T^\prime, A^\prime)=(T,A) }} p(\varepsilon|\mathcal{I};\theta)
%                         \sum\limits_{\substack{E^\prime \in \mathbb{E}_{\Sigma}^*  \\ E^\prime(\text{``''},y_{1:0})=(T^\prime,A^\prime)}} p(E^\prime|\mathcal{I};\theta) \\
%               =& \sum\limits_{\substack{\varepsilon \in \mathbb{E}_\Sigma^*  \\   (T^\prime, A^\prime) \in (\mathbb{T}_\Sigma, \mathbb{P}_{\Sigma}^*) \\ \varepsilon(T^\prime, A^\prime)=(T,A) }} p(\varepsilon|\mathcal{I};\theta) \ep(T^\prime, A^\prime)\\
               =&\left\{
               	        \begin{array}{lr}
               		        \multicolumn2l{1 \hspace{34mm} \mif i=0, j=0}\\
%               		        {\ep(T_{1:i},A_{1:j-1})R_j(2)} &
%               		        {\mif i=0, j>0} \\
%               		        {\ep(T_{1:i-1},A_{1:j})R_j(3)I_j(T_i)}&
%               		        {\mif i>0, j=0} \\
%               		        {\ep(T_{1:i},A_{1:j-1})R_{j+1}(2)}&{\mif i>0, j>0,} \\
%               		         \multicolumn2l{~~ +\ep(T_{1:i-1},A_{1:j})R_j(3)I_j(T_i)~~~\text{and~}T_{i}\ne\text{`\#'}}\\
%               		         \multicolumn2l{~~ +\ep(T_{1:i-1},A_{1:j-1})R_j(1)A_j(T_i) } \\
%               		         {\ep(T_{1:i},A_{1:j-1})}  & {\mif i>0, j>0,} \\
%               		         \multicolumn2l{~~ +\ep(T_{1:i-1},A_{1:j})R_j(3)I_j(T_i)~~~\text{and~}T_{i}=\text{`\#'}}\\
%               		         \multicolumn2l{~~ +\ep(T_{1:i-1},A_{1:j-1})R_j(1)A_j(T_i) } \text{.}\\
               		         p(\varepsilon^D_{i,j}|\mathcal{I};\theta)\ep(T_{1:i},y_{1:j-1})\delta_{j>0} & \text{otherwise}\\
               		         \multicolumn2l{~~ + p(\varepsilon^I_{i,j}|\mathcal{I};\theta)\ep(T_{1:i-1},y_{1:j})\delta_{i>0}}\\
               		         \multicolumn2l{~~ + p(\varepsilon^C_{i,j}|\mathcal{I};\theta)\ep(T_{1:i-1},y_{1:j-1})\delta_{i>0,j>0}}
               	        \end{array}
               	        \right.
		        %\text{.}
             \end{array}
        \end{equation}
\endgroup
where the value of $\delta_{condition}$ is $1$ if the condition is met, otherwise $0$.
By recursively applying Eq.~\ref{eq:eprewrite}, $\EP(T|\mathcal{I};\theta) = \ep(T,y)$ can be calculated in $\bigO(|T|\cdot|y|)$.

Fig.~\ref{fig:ep} shows the EP calculation process for generating ``DOVE\#'' from the sequence of \emph{pd} displayed in Fig.~\ref{fig:mot}~(a). We can see that the insertion operation, which inserts an `O' is contained in the maximal probable edit path. This is an expected result.

\subsection{EP Training}
With the training set $\mathcal{X}$ that consists of pairs of image and \emph{gt} string, the EP training is to find $\hat{\theta}$ that minimizes the negative log-likelihood over $\mathcal{X}$:
%The aim of maximum likelihood training is to minimize the negative log-probabilities of all the ground truth sequence conditioned on the corresponding input data in the training set. For a particular training image $\mathcal{I}$, that is, to minimize $\p(T', A)$ where the character sequence $T'$ is the ground truth and the sequence of probability distribution $A$ is the output of the sequence-recognition-based neural network with input $\mathcal{I}$.
%objective function:
\begin{equation}\label{epl}
\hat{\theta}=\arg\min_{\theta}-\sum_{(\mathcal{I},T)\in\mathcal{X}}\ln(\EP(T|\mathcal{I};\theta))
\end{equation}
%where $A$ is the output sequence of \emph{pd} conditioned on $\mathcal{I}$ under model parameters $\theta$ and is computed by Eq.~\ref{eq:att}, while $\ep(T^\prime, A)$ is the edit probability of edit state $(T^\prime, A)$ defined in Eq.~\ref{eq:ep} and computed by Eq.~\ref{eq:eprewrite}.
%where $T^\prime$ is the ground truth.
The model can be optimized with standard back-propagation algorithm~\cite{rumelhart1988learning}.

\subsection{EP Predicting}\label{sec:epp}
EP Predicting is to find the string $\hat{T}$ that maximizes $\EP(\hat{T}|\mathcal{I};\hat{\theta})$:
%where A is the output sequence of \emph{pd} conditioned on the input data $\mathcal{I}$:%, that is, to find the prefix sequence $\hat{T}\oplus EOS$ which equals to maximize the probability conditioned on the distribution sequence $A$:
    	\begin{equation}\label{eq:epp}
    		\hat{T} = \arg\max_{T\in{\mathbb{T}_\Sigma}}\EP(T|\mathcal{I};\hat{\theta})\text{.}
    	\end{equation}
However, looking for the whole answer set ${\mathbb{T}_\Sigma}$ with Eq. \ref{eq:epp} is extremely costly.
Therefore, we develop two efficient sequence generation mechanisms for both lexicon-free and lexicon-based prediction.
    	
\textbf{Predicting without lexicon.}
%In general, for a well trained model,  the character sequences with the largest edit probability are more probably generated by directly consuming each probability of $A$.
%Therefore, instead of looking for the whole answer set $\Sigma^*$, we define a candidate set $\mathcal{B}_A$ based on $A$ as following:
 %   	\begin{equation}\label{eq:cand}
  %          \mathcal{B}_A=\pref(\max A_1 \oplus \max A_2  \oplus \cdots  \oplus A_{L-1})
   % 		%\mathcal{B}_A=\pref(\sum_{i=1}^{L-1}\max_{k\in\Sigma-EOS}A_i(k))
%    	\end{equation}
%where $\pref(S)$ means the set of $S$'s prefixes.
%For example,  given $A$ in Fig. \ref{fig:mot}$ (a), \mathcal{B}_A$ consists of ``'', ``A'', ``AC'', ``ACD'', ``ACDD'' and ``ACDDD''.
%Note that, the $EOS$ ($\#$) is removed.
%For each $\tilde{T}\in\mathcal{B}_A$, $\ep(\tilde{T} \oplus EOS, A)$ is lexicographically calculated in O($L$) with Eq. \ref{eq:ep}.
%Finally, we select the one with the maximal probability generated from $A$ as the predicted result:
%    	\begin{equation}\label{eq:eplp}
%    		\hat{T} = \max_{\tilde{T}\in\mathcal{B}_A}\ep(\tilde{T} \oplus EOS, A)\text{.}
 %   	\end{equation}
 %After carefully observation and analysis
 By deeply analyzing the prediction problem, we find that in general,
 %{\color{red}
 the string $\hat{T}$ that maximizes $\EP(\hat{T}|\mathcal{I};\hat{\theta})$ is mostly a prefix (ended by an EOS `\#') of the string $\mathcal{T}$ with the most probable edit path that transforms $(\text{``''}, y_{1:0})$ to $(\mathcal{T}, y)$ where $\text{`\#'}\not\in\mathcal{T}$.

 Therefore, we firstly find the string $\mathcal{T}$:
  \begin{equation}\label{eq:T}
 %\arg\max_{\tilde{T}\in\Sigma^*}\ep(\tilde{T}+EOS, A) =
 	\mathcal{T} = \arg\max_{{T}\in(\Sigma\backslash\{\text{`\#'}\})^*}\max_{\substack{E\in\mathbb{E}_{\Sigma}^*\\E(\text{``''}, y_{1:0})=({T}, y)}}p(E|\mathcal{I};\theta)
 \end{equation}
 where $\Sigma\backslash\{\text{`\#'}\}$ represents the alphabet without the EOS,
 then use all the prefixes of $\mathcal{T}$ (each ended by an EOS) as the candidate set $\mathcal{B}$:
 \begin{equation}\label{eq:cand}
	\mathcal{B} = \{\mathcal{T}_{1:i}\oplus\text{`\#'}|0\le i\le|\mathcal{T}|\}
 \end{equation}
 where $\oplus$ represents the concatenation operator for two strings,
 finally select the best $\hat{T}\in\mathcal{B}$ that maximizes $\ep(\hat{T}, A)$:
  \begin{equation}\label{eq:cand}
 	\hat{T} = \arg\max_{{T}\in\mathcal{B}}\EP({T}|\mathcal{I};\hat{\theta})\text{.}
 \end{equation}

 %{\color{red}
Note that the edit path with the highest probability should not include an insertion edit operation that inserts a non-EOS character,
because
%$(E_2\circ\varepsilon^I_{i,j}\circ E_1)(y_{1:0},y_{1:0})=(T_1\oplus T_2, A) \Rightarrow(E_2\circ E_2)(y_{1:0},y_{1:0})=(T_1\oplus T_2, A)$
%while
%$p(E_2\circ\varepsilon_I^k\circ E_2) = p(E_2\circ E_2|\mathcal{I};\theta)p(\varepsilon^I_{i,j}|\mathcal{I};\theta) \le p(E_2\circ E_2|\mathcal{I};\theta)$.
we can remove such insertions and get a new edit path whose conditional probability is greater than the previous one.

Therefore, we can generate $\mathcal{T}$ by beginning with the state $(\text{``''},y_{1:0})$ and performing the most probable deletion or consumption operation~(not considering any operation generating the EOS) at each step, till a state $(\mathcal{T}, y)$ is reached. Since all strings in $\mathcal{B}$ share the common prefixes, we can compute all the $\ep({T}, y)$ for ${T}\in\mathcal{B}$ in $O(|\mathcal{T}|\cdot|y|)$ time with Eq.~\ref{eq:eprewrite}.
%\begin{equation}
%\begin{array}{lcr}
%	\mathcal{T} &\defeq& \mathcal{T}_{A}\\
%	\mathcal{T}_{A_{1:j}}&=&\left\{\begin{array}{lr}
%		\mathcal{T}_{A_{1:j}} &\\
%		\mathcal{T}_{A_{1:j}}\oplus &
%	\end{array}\right.\\
%	\mathcal{T}_{A_{1:0}}&=&\text{``''}
%\end{array}
%\end{equation}
%from state $(\text{``''}, y_{1:0})$, and on each state $(T_{1:i}, A_{1:j})$, we separately perform:
%\begin{itemize}
%	\item ending the string with appending an $EOS$ directly or by consuming the $A_{j+1}$(if $j<|A|$), then add the string into the candidate set $\mathcal{B}_A$;
%	\item select the highest possible edit operation from deletion or consumptions (generating any character except for $\#$ by consuming the next probability distribution), then perform it.
%	%}
%\end{itemize}
%Finally, we select the one with the maximal edit probability generated from $A$ as the predicted result:
%    \begin{equation}\label{eq:eplp}
%    		\hat{T} = \arg\max_{\tilde{T}\in\mathcal{B}_A}\ep(\tilde{T} \oplus EOS, A)\text{.}
%   	\end{equation}

\begin{figure}[!thbp]
	\begin{center}
		\includegraphics[width=0.45\textwidth]{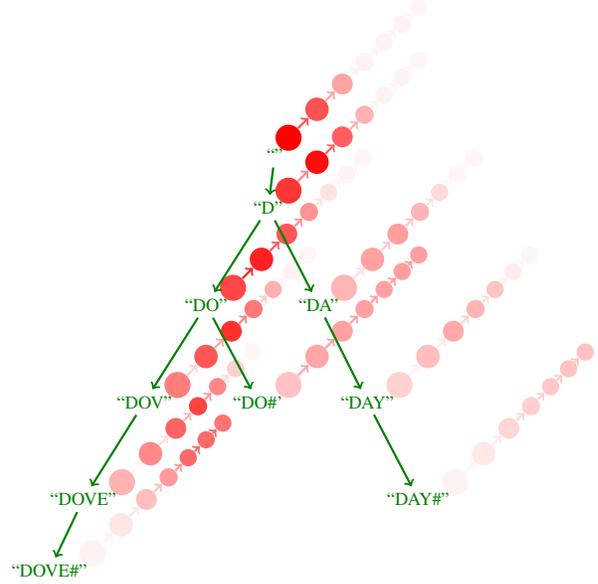}
	\end{center}
	\caption{The illustration of edit probability trie.
		The trie contains a lexicon with three strings: ``DOVE\#'', ``DO\#'' and ``DAY\#''.
		Each node represents a prefix of one or more strings in the lexicon.
		Upper right to each node with the prefix (say $S$) is a probability vector where the color saturation of the $j$-\emph{th} element represents $\ep(S, y_{1:j})$ and $y$ is the output sequence of \emph{pd}.
		Higher color saturation means higher probability, and vice versa.
	}
	\label{fig:eptrie}
\end{figure}

\textbf{Predicting with a lexicon.}\label{sec:pwl}
In constrained cases, we can enumerate the strings in a lexicon $\mathcal{D}$, and find the most probable one.
A tunable parameter $\lambda$ is used to indicate how much we trust the lexicon $\mathcal{D}$ since some target strings may be not contained in the lexicon.
Therefore, we actually assess all the possible strings in $\mathcal{B} \cup \mathcal{D}$ by
    	\begin{equation}
    		\hat{T} = \arg\max_{{T}\in\mathcal{B}\cup\mathcal{D}}\left\{
    		\begin{array}{lr}
    			\lambda\EP({T}|\mathcal{I};\hat{\theta}) & \mif {T}\in\mathcal{D}\\
    			(1-\lambda)\EP({T}|\mathcal{I};\hat{\theta}) & \mif {T}\not\in\mathcal{D}
    		\end{array}
    		\right.
    	\end{equation}
where $0.5\le\lambda\le1$.
The larger $\lambda$ is, the more we trust the lexicon, and vice versa.
Specifically, $\lambda=0.5$ means that the lexicon can provide only some additional candidates that are treated equally to those in $\mathcal{B}$, while $\lambda=1$ means that the generated strings are guaranteed to appear in the lexicon $\mathcal{D}$.
	
However, with the growing of lexicon size, the above enumeration-based method is extremely time-consuming.
To tackle this problem, Shi \etal used a prefix tree (Trie)~\cite{briandais1959trie,shi2016robust} to accelerate the search process since many strings in the lexicon share common prefixes.
In this work we develop a data structure called \textbf{edit probability Trie (EP-Trie)}, which is a variant of Trie with nodes containing not only a prefix $S$,
but also a vector indicating $\ep(S, y_{1:j})$ for $0\le j\le|y|$.
The vector of a node can be computed from the vector of its parent in O($|y|$) time with Eq.~\ref{eq:eprewrite}.
We will demonstrate the effectiveness of EP-Trie in Sec.~\ref{sec:dop}. Fig.~\ref{fig:eptrie} illustrates EP-Trie.

%------------------------------------------------------------------------
\begin{table*}[!htbp]
	\begin{center}
		\begin{tabular}{|l||c|c|c||c|c||c|c|c||c||c|}
			%\centering
			\hline
			\multirow{2}{*}{\textbf{Method}} &
			\multicolumn{3}{c||}{\textbf{IIIT5k}} & \multicolumn{2}{c||}{\textbf{SVT}} & \multicolumn{3}{c||}{\textbf{IC03}} & \textbf{IC13} & \textbf{IC15} \cr\cline{2-11}
			& \textbf{50} & \textbf{1k} & \textbf{None} & \textbf{50} & \textbf{None} & \textbf{50} & \textbf{Full} & \textbf{None} & \textbf{None} & \textbf{None} \cr\hline
			ABBYY~\cite{wang2011end} & 24.3          & $-$ & $-$ & 35.0 & $-$ & 56.0 & 55.0 & $-$ & $-$  & $-$\cr
			Wang \etal\cite{wang2011end}           & $-$  & $-$  & $-$ & 57.0 & $-$ & 76.0 & 62.0 & $-$ & $-$  & $-$\cr
			Mishra \etal\cite{graves2013speech}         & 64.1 & 57.5 & $-$ & 73.2 & $-$ & 81.8 & 67.8 & $-$ & $-$  & $-$\cr
			Wang \etal\cite{wang2012end}           &$-$  & $-$  & $-$ & 70.0 & $-$ & 90.0 & 84.0 & $-$ & $-$  & $-$\cr
			Goel \etal\cite{goel2013whole}           &$-$  & $-$  & $-$ & 77.3& $-$ & 89.7& $-$  & $-$ & $-$  & $-$\cr
			Bissacco \etal\cite{bissacco2013photoocr}       &$-$  & $-$  & $-$ & 90.4& 78.0& $-$ & $-$ & $-$ & 87.6 & $-$\cr
			Alsharif and Pineau~\cite{alsharif2013end}   &$-$  & $-$  & $-$ & 74.3& $-$ & 93.1 & 88.6 & $-$ & $-$  & $-$\cr
			Almaz{\'a}n \etal\cite{almazan2014word}         &91.2  & 82.1 & $-$ & 89.2 & $-$ & $-$ & $-$ & $-$ & $-$  & $-$\cr
			Yao \etal\cite{yao2014strokelets}            &80.2 & 69.3 & $-$ & 75.9& $-$ & 88.5 & 80.3 & $-$ & $-$  & $-$\cr
			Rodr{\'i}guez-Serrano \etal\cite{rodriguez2015label}  &76.1 & 57.4 & $-$ & 70.0& $-$ & $-$ & $-$ &  $-$ & $-$  & $-$\cr
			Jaderberg \etal\cite{jaderberg2014deep}      &$-$  & $-$  & $-$ & 86.1& $-$ & 96.2& 91.5 & $-$ & $-$  & $-$\cr
			Su and Lu~\cite{su2014accurate}             &$-$  & $-$  & $-$ & 83.0& $-$ & 92.0& 82.0 & $-$ & $-$  & $-$\cr
			Gordo~\cite{gordo2015supervised}                 &93.3 & 86.6 & $-$ & 91.8& $-$ & $-$ & $-$ & $-$ & $-$  & $-$\cr
			Jaderberg \etal\cite{jaderberg2016reading}      &97.1 & 92.7 & $-$ & 95.4& 80.7& 98.7& \textbf{98.6}& 93.1& 90.8 & $-$\cr
			Jaderberg \etal\cite{jaderberg2014deep}      &95.5 & 89.6 & $-$ & 93.2& 71.7& 97.8& 97.0& 89.6& 81.8 & $-$\cr
			Shi \etal\cite{shi17end}            &97.6 & 94.4 & 78.2& 96.4& 80.8& 98.7& 97.6& 89.4& 86.7 & $-$\cr
			Shi \etal\cite{shi2016robust}            &96.2 & 93.8 & 81.9& 95.5& 81.9& 98.3& 96.2&  90.1& 88.6 & $-$\cr
			Lee \etal\cite{lee16recursive}    &96.8 & 94.4 & 78.4& 96.3& 80.7& 97.9& 97.0&  88.7& 90.0 & $-$\cr
			Cheng \etal\cite{cheng17fan}       &99.3 & 97.5 & 87.4& \textbf{97.1}& 85.9& \textbf{99.2}& 97.3&  {94.2}& {93.3} & 70.6\cr\hline
			%Shi's + CE        &98.5 & 96.9 & 83.8& 94.4& 79.9& 98.6& 96.7&  90.5& 89.2 & 62.3\cr
			Shi \etal (baseline) \cite{shi2016robust}            &96.5 & 92.8 & 79.7& 96.1& 81.5& 97.8& 96.4&  88.7& 87.5 & $-$\cr
			Cheng \etal (baseline) \cite{cheng17fan}       &98.9 & 96.8 & 83.7& 95.7& 82.2& 98.5& 96.7&  91.5& 89.4 & 63.3\cr\hline
			Shi's + EP (ours) & {99.1} & {97.3} & {85.0}& {96.3}& {86.2}& {98.4}& 97.0& {93.7}& {93.0} & {68.1} \cr
			%Cheng's + CE        &\textbf{99.3} & \textbf{97.9} & 87.3& \textbf{96.8}& \textbf{87.5}& \textbf{99.1}& {98.2}&  \textbf{94.0}& \textbf{93.2} & \textbf{71.7}\cr
			Cheng's + EP (ours) &\textbf{99.5} & \textbf{97.9} & \textbf{88.3}& {96.6}& \textbf{87.5}& {98.7}& {97.9}& \textbf{94.6}& \textbf{94.4} & \textbf{73.9} \cr\hline
		\end{tabular}
	\end{center}
	\caption{Results of recognition accuracy on general benchmarks. ``50'' and ``1k'' are lexicon sizes, ``Full'' indicates the combined lexicon of all images in the benchmarks, and ``None'' means lexicon-free. The results of the baseline methods are directly referenced from the ``SRN only'' and the ``Baseline'' in \cite{shi2016robust} and \cite{cheng17fan} respectively.}
	\label{tab:results}
\end{table*}
\section{Performance Evaluation}\label{sec:performance}
We conduct extensive experiments to validate the EP method on several general recognition benchmarks under the attention framework.
For a fair and comprehensive comparison,
% in addition to existing methods,
%we also evaluate the state-of-the-art models with a CE Loss layer,
%which is taken as the baseline method.
% first
we directly employ the structures of the state-of-the-art works and replace their loss layers with EP.
We first evaluate the performance of the EP-based methods, and compare them with the existing methods.
%second
Then we demonstrate the advantage of EP training over frame-wise loss based training on some real training data.
Finally, we evaluate our method with the Hunspell 50k lexicon~\cite{hunspell}, and compare EP predicting methods with and without lexicon. %discuss the effectiveness of the proposed EP predicting methods.}

\begin{figure*}[!thbp]
	\begin{center}
		\includegraphics[width=\textwidth]{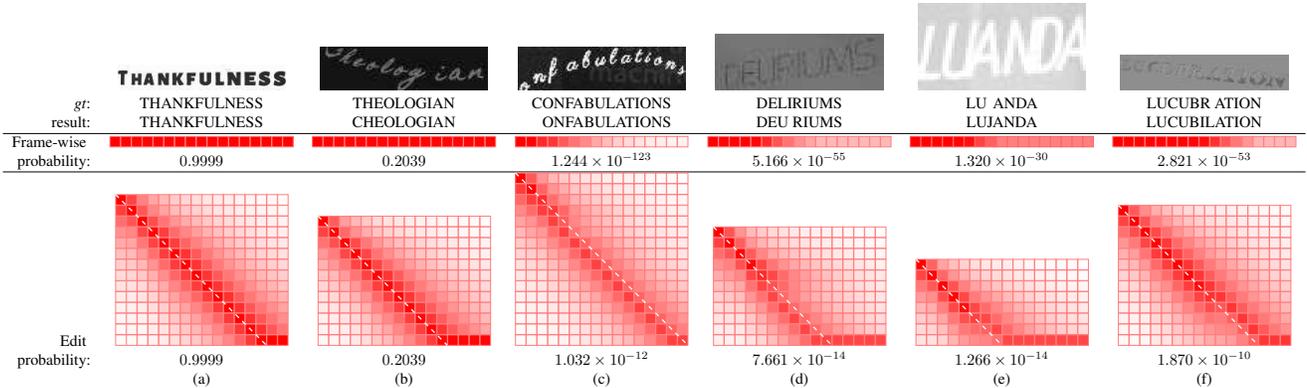}
	\end{center}
	\caption{The visual comparison of EP and FP on real training data.
		In each subfigure, the characters shown in the 3rd raw are those having the largest probability in the corresponding \emph{pd}.
		A vector of probability is shown in the 4th row, where the red saturation of the $j$-\emph{th} element (labeled from $0$) of the vector indicates the FP value of the first $j$ characters in the \emph{gt}, calculated by the first $j$ \emph{pd}s in the output sequence.
		A matrix of probability is given in the 6th row, where the red saturation of the ($i$, $j$) element of the matrix indicates the EP value of generating the first $i$ characters in the \emph{gt} from the first $j$ \emph{pd}s in the output sequence.
		%To facilitate observation, a white dashed line is displayed on the diagnoses of each edit probability matrix.
		In subfigure (a) and (b), no misalignment occurs. The maximal possible edit path appears on the diagonal line of the EP matrix because no insertion/deletion operation is likely to be performed. %Therefore, the value of the edit probability is almost equal to the frame-wise likelihood's.
		In subfigure (c) and (d), some misalignments occur because some characters are missing. The maximal possible edit path appears below the diagonal line of the EP matrix because some insertions are likely to be performed near by the places of the missing characters.
		In subfigure (e) and (f), some misalignments occur because some characters are superfluous. The maximal possible edit path appears to the right of the diagonal line of the EP matrix because some deletions are likely to be performed near by the places of the superfluous characters.
		%And the value of EP loss is significantly smaller than FW loss's when misalignments occur.
	}
	\label{fig:demo}
\end{figure*}

    \subsection{Datasets}
		{\bf{IIIT 5K-Words}} \cite{mishra2012scene} (IIIT5K) is a dataset collected from the Internet with 3000 cropped word images in its test set. For each of its images, a 50-word lexicon and a 1k-word lexicon are specified, both of which contain the ground truth words as well as other randomly picked words.
		
		{\bf{Street View Text}} \cite{wang2011end} (SVT) is collected from the Google Street View. Its test set consists of 647 word images, each of which is specified with a 50-word lexicon.
		
		{\bf{ICDAR 2003}} \cite{lucas2003icdar} (IC03) contains 251 scene images with text bounding boxes. Each image is associated with a 50-word lexicon defined by Wang \etal \cite{wang2011end}. A full lexicon that combines all lexicon words is also provided. For fair comparison,  as in \cite{wang2011end}, we discard the images containing non-alphanumeric characters or have less than three characters. The resulting dataset contains 867 cropped images.
		
		{\bf{ICDAR 2013}} \cite{karatzas2013icdar} (IC13) is the successor of IC03, so most of its data are inherited from IC03. It contains 1015 cropped text images, but no lexicon is associated.
		
		{\bf{ICDAR 2015}} \cite{karatzas2015icdar} (IC15) contains 2077 cropped images. For fair comparison, we discard the images containing non-alphanumeric characters, and eventually obtain 1811 images in total. No lexicon is associated.

\subsection{Implementation Details}
\textbf{Network Structure:}
%    	\begin{figure}[!thbp]
%    		\begin{center}
%    			\includegraphics[width=0.35\textwidth]{images/struct.eps}
%    		\end{center}
%    		\caption{The structure of the EP-Component.
%                            A sequence-recognition-based neural network $f$ first transform input data $\mathcal{I}$ to a sequence of probability distribution  $A$.
%                            Two trainable fully connect functions $WA+\omega$ and $UA+\mu$ are separately exploited for calculating $P_w$ and $I$, and $\sigma$ means the softmax function.
%                            %Then, trainable functions $g$ and $h$ calculate $P_W$ and $P_{ins}$ according to $A$ respectively.
%                            In the training phase, the component calculates the negative log-probability $J$ of ground truth $T^\prime$ conditioned on $\mathcal{I}$ with insertion and deletion taken into account, then conducts the errors' back-propagation;
%                            In the predicting phase, the component finds the best result $\hat{T}$ in candidate set $\mathcal{B} \cup D$ by maximizing $\ep(T, f(\mathcal{I}))$.
%                            }
%    		\label{fig:str}
%    	\end{figure}
%The-state-of-the-art for character sequence generation  is attention-based encoder/decoder framework,
The attention-based encoder-decoder framework is the-state-of-the-art technique for text recognition, which consists of two major steps:
1) Obtaining visual feature representation with a CNN-based feature extractor, such as 7-Conv-based by Shi \etal\cite{shi2016robust} and ResNet-based by Cheng \etal\cite{cheng17fan};
2) Generating the output sequence of probability distribution with the attention model.
In this work, we %first test both of the Shi's and Cheng's structures with CE as the baseline,
%then
evaluate the proposed method by replacing the FP-based training/predicting in Shi's and Cheng's structures with the EP-based training/predicting.

%7-Conv-based feature extractor proposed by Shi \etal or a ResNet-based feature extractor proposed by Cheng \etal. In this work, we reproduced the above two structures with a CE Loss layer as the baseline methods. And then we replace the loss layer with the EP-Component to test the effect of the proposed method. The EP-Component is shown in Fig \ref{fig:str}. In this implementation, we use full connected layer as function $g$ and $h$.

        \textbf{Model Training:} Our model is trained on 8-million synthetic data released by Jaderberg~\etal \cite{jaderberg2014synthetic} and 4-million synthetic data (excluding the images that contain non-alphanumeric characters) cropped from 800-thousand images released by Gupta \etal~\cite{Gupta16} by the ADADELTA~\cite{adadelta} optimization method.

        %We set the batch size as 150 and scale images to 256 × 32 in both training and predicting.
        %Our model processes about 250 samples per second, and converges in 4 days after about 8 epochs over the training set.
    	
    	\textbf{Running Environment:} Our method is implemented under the Caffe framework~\cite{jia2014caffe}. In our implementation, most modules in our model can be GPU accelerated as the CUDA backend is extensively used. All experiments are conducted on a workstation equipped with an Intel Xeon(R) E5-2650 2.30GHz CPU, an NVIDIA Tesla P40 GPU and 128GB RAM.

\subsection{Comparison with Existing Methods}
Tab.~\ref{tab:results} shows the performance results of two EP-based methods, two FP-based (baseline) methods and previous methods.
%We first compare the results of EP-based methods and the baselines with Shi's and Cheng's structure,
With Shi's and Cheng's structures, the EP-based methods significantly outperform the baseline methods on all benchmarks.
In comparison with the existing methods, we consider both constrained and unconstrained cases.
In the  unconstrained cases, the EP-based method (Cheng's + EP) outperforms all existing methods;
While in the constrained cases, our method (Cheng's + EP) performs better than all existing methods on IIIT5K, and achieves comparable results to that of the method proposed by Cheng \etal (FAN) \cite{cheng17fan} on SVT and IC03 datasets.
However, it should be pointed out that FAN is trained with both word-level and character-level bounding box annotations, which is expensive and time consuming. On the contrary, our method is trained with only word-level ground truth.
We also note that the method proposed by Jaderberg \etal \cite{jaderberg2014synthetic} achieves the best result on IC03 with the full lexicon, but it cannot recognize out-of-training-set words.

%To compare with Cheng \etal's model in \cite{2017arXiv170902054C},
%In order to further demonstrate the performance of our method, we also evaluate the total normalized edit distance (NED)\cite{karatzas2013icdar} on all benchmarks for the baseline and EP-based methods, shown in Tab. \ref{tab:ned}.
%We find that the EP-based method Cheng's+Ep significantly outperforms the results of baseline in all datasets, and achieve the competitive performance comparing to the results on IIIT5k with 50 lexicon, SVT with 50 lexicon and IC03 with 50 lexicon.
%Tab. \ref{tab:ned} shows the results
%, and the results are shown in Tab. \ref{tab:ned}. We find that EP Net significantly improves the NED measure in both constrained and unconstrained cases by comparing to the baseline and Cheng's result.
%\subsection{The Effect of Ground-Truth-Unrelated Lexicon}\label{sec:aoe}

\subsection{Performance of EP Training}\label{sec:dot}
To demonstrate the advantage of EP in training stage, we compare the calculation of the FP and the proposed EP on some real training data. The input images, the ground truths and the recognition results are displayed in the 1st, 2nd and 3rd rows in the upper part of Fig.~\ref{fig:demo}, respectively.
As for the recognition results, they are actually the \emph{pd} sequences. For demonstration convenience, we just display the characters that dominate the corresponding \emph{pd}.

For the FP calculation, we show a vector of probability for each image on the 4th row and the FP value on the 5th row in Fig.~\ref{fig:demo}.
The value of the $j$-\emph{th} element in the vector represents the joint conditional probability of generating the first $j$ characters in \emph{gt} from the first $j$ \emph{pd}s in the output $pd$ sequence, and the probabilities after the EOS are ignored (regarded as $1$).
We have the following observations on FP:
1) when the output \emph{pd} sequence is well aligned to the \emph{gt} (see Fig.~\ref{fig:demo} (a) and (b)), the FP declines only at the place where the character is misclassified;
2) when the output \emph{pd} sequence is misaligned to the \emph{gt} (see Fig.~\ref{fig:demo} (c), (d), (e) and (f)), the FP continues to decline after any occurrence of missing/superfluous character even some characters are correctly recognized.
As a result, in the cases of misalignment, the error will be back-propagated to the correctly recognized \emph{pd}s following the missing/superfluous character, which may confuse the model training.

For the EP calculation, we display a matrix of probability for each image on the 6th row and the EP values on the 7th row. The value of the ($i$, $j$) element represents $\ep(T_{1:i}, y_{1:j})$ where $T$ is the \emph{gt} and $y$ is the output sequence of \emph{pd} conditioned on the input image $\mathcal{I}$.
We have the following observations on EP:
1) When the output sequence of \emph{pd} is well aligned to the \emph{gt} (see Fig.~\ref{fig:demo} (a) and (b)), no matter whether the classification result is correct, the EP value is almost equal to the FP value and the most probable edit path appears on the matrix diagonal line, that is, generating every character in \emph{gt} by consuming the corresponding \emph{pd};
2) When some characters are missing/superfluous (see Fig.~\ref{fig:demo} (c) and (d) for missing character cases, (e) and (f) for superfluous character cases), the EP value is much larger than the FP value, and the most probable edit path appears under or to the right of the matrix diagonal line after the occurrence of the missing/superfluous characters. Different from the FP, the EP is indicative of inserting/deleting the missing/superfluous character and generating others by consuming the aligned \emph{pd}.
As a result, in the cases of misalignment, much error will be back-propagated only to the place where the missing/superfluous character occurs, and little error back-propagated to the correctly recognized \emph{pd}s before or after the occurrence of the missing/superfluous characters, which makes the model training process focus on the missing/superfluous characters, instead of the misaligned characters.

As  EP-based methods theoretically require more calculation than baselines, we measure the time cost for training. The result shows that Shi's/Cheng's baselines cost 170.5ms/536.0ms per iteration, and EP based methods cost only 6.8ms/7.0ms more (batch size is set to 75).

\subsection{Performance of EP Prediction}\label{sec:dop}
\begin{figure}[!thbp]
   	\begin{center}
   		\includegraphics[width=0.48\textwidth]{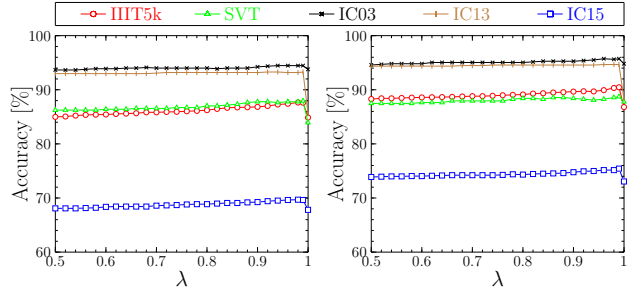}
   	\end{center}
   	\caption{The accuracy on general recognition datasets when predict with the Hunspell 50k lexicon and $\lambda$ varies from 0.5 to 1.
 The left and right figures are the results of the Shi's+EP method and Cheng's+EP method, respectively.}
   	\label{fig:g}
\end{figure}

Here, we evaluate the performance of EP prediction with/without lexicon.
In real world text recognition tasks, it is not easy to get the ground-truth-related lexicons.
Therefore, following previous works~\cite{alsharif2013end,jaderberg2014deep,jaderberg2016reading,shi17end,shi2016robust}, we also test our methods on a public ground-truth-unrelated lexicon Hunspell~\cite{hunspell}, which contains more than 50k words.
% , containing 79672 English words and provided by The English Lexicon Project\cite{Balota2007}.
%Results listed in Table \ref{tab:results}  are based on none lexicon or ground-truth-based lexicons.
%Some results with ground-truth-related lexicons are listed in Tab.~\ref{tab:results}.
%Results listed in Table \ref{tab:results}  are based on none lexicon or ground-truth-related lexicons. However, in real scene text reading task, we can not get the ground truth previously to generate the ground-truth-related lexicon. So, we also conduct experiments on a ground-truth-unrelated lexicon which contains 79672 English words, provided by English Lexicon Program.
As mentioned in Sec.~\ref{sec:pwl}, $\lambda$ is a tunable super-parameter in the predicting stage.
We conduct lexicon-based prediction on all datasets by varying $\lambda$ from $0.5$ to $1$.
The results are shown in Fig.~\ref{fig:g}. We can see that 1) the accuracy increases when $\lambda$ varies from $0.5$ to $0.98$, but
%We test different $\lambda$ for $\frac{\lambda}{1+\lambda}\in[0.5,1)$ on the general recognition datasets.
decreases rapidly when $\lambda$ approaches 1 due to over-correction;
%In fact, the effect of $\lambda$ may be different on different datasets
%Besides, after amount of experiments, we find that the effect of $\lambda$ is slightly different on different datasets.
%, but a value about $0.95\sim0.98$ should be always not bad.
    	%To fairly compare, we also redress the result of the baselines methods with the same lexicon by the following function:
    	%\begin{equation}
    	%	\hat{T} = \min_{T'\in\mathcal{D}}\ed(T', T_0)
    	%\end{equation}
    	%where $T_0$ is the prediction result of the frame-wise prediction.
%Tab. \ref{tab:pwla}  show the accuracy of the EP-based methods with Hunspell.
2) When $\lambda$ is set to $0.5$, the accuracy of lexicon-based prediction is exactly the same as that of lexicon-free prediction, which demonstrates the effectiveness of the proposed lexicon-free prediction method;
3) The ground-truth-unrelated lexicon is also helpful for improving text recognition performance if $\lambda$ falls in a proper range (from $0.9$ to $0.98$), which validates the effectiveness of the lexicon-based prediction method.
    	
%We carry out the character sequence
%In predicting stage, we test the enumeration-based and EP-Trie-based methods on the text recognition results and time consumption of the whole 5 tested datasets.
%Experiments show that 1) the results of EP-Trie is same to the enumeration's, and 2) EP-Trie ends in 235 seconds while enumeration ends in 33782 seconds, which demonstrate the excellent performance of EP-Trie.
Besides,
we also test the enumeration-based and EP-Trie-based methods in terms of recognition accuracy and time cost per image while using the 50k lexicon.
Our experiments show that 1) the EP-Trie-based method predicts the same result as the enumeration-based method's, and 2) the former costs 0.11 second per image while the latter costs 2.566 seconds per image, which demonstrates the excellent efficiency of EP-Trie.

%We implement prediction method with both enumeration and EP-Trie, and test the time consumptions of them.
%They predict the same result.
%And the time consumptions listed on Table \ref{tab:prtrie} show that EP-Trie runs extremely faster than enumeration.

%------------------------------------------------------------------------
\section{Conclusion}\label{sec:conclusion}
In this work, we propose a new method called edit probability for accurate scene text recognition. The new method can effectively handle the misalignment problem between the training text and the output probability distribution sequence caused by missing or superfluous characters. %method learning dislocation problem existed in the existing objective functions,
%then develop the edit probability to solve the problem,
%finally we also design the corresponded training/predicting methods for model training/predicting.
We conduct extensive experiments over several benchmarks to validate the proposed method, and experimental results show that EP can significantly improve recognition performance. %outperforms existing methods.
%And as EP-Component does not need extra training data or labels, it makes a great performance improvement with a very small cost.
In the future, we plan to apply the EP idea to machine translation, speech recognition, image/video caption and other related tasks.

%------------------------------------------------------------------------

\section{Acknowledgment}\label{sec:ack}
Fan Bai and Shuigeng Zhou were partially supported by the Science and Technology Innovation Action Program of Science and Technology Commission of Shanghai Municipality (STCSM) under grant No.~17511105204.

{\small
\bibliographystyle{ieee}
\bibliography{egbib}
}

\end{document}